\documentclass[twoside,11pt]{article}

\usepackage[arxiv]{melba}

\usepackage{amsmath,amsfonts}

\usepackage{framed}
\usepackage{xcolor}

\newcommand\crule[1][black]{\textcolor{#1}{\rule{0.7em}{0.7em}}}

\melbaid{YYYY:NNN}  
\melbaauthors{Beliveau et al.}  
\volume{1}
\firstpageno{1}  
\melbayear{2024}  
\datesubmitted{08/2024}  

\melbaspecialissue{Medical Imaging with Deep Learning (MIDL) 2020}
\melbaspecialissueeditors{Marleen de Bruijne, Tal Arbel, Ismail Ben Ayed, Hervé Lombaert}

\ShortHeadings{Classification of Radiological Text}{Beliveau et al.}

\title{Classification of Radiological Text in Small and Imbalanced Datasets in a Non-English Language}

\author{\name Vincent Beliveau \email{vincent.beliveau@nru.dk} \AND
\addr Neurobiology Research Unit, Rigshospitalet, Copenhagen, Denmark
\AND
\addr Department of Neurology, Medical University of Innsbruck, Innsbruck, Austria \AND
\name Helene Kaas \email{helene.kaas@nru.dk} \AND
\addr Neurobiology Research Unit, Rigshospitalet, Copenhagen, Denmark \AND
\addr Epilepsy Clinic, Department of Neurology, Rigshospitalet, Copenhagen, Denmark \AND
\name Martin Prener \email{martin.prener@nru.dk} \AND
\addr Neurobiology Research Unit, Rigshospitalet, Copenhagen, Denmark \AND
\addr Epilepsy Clinic, Department of Neurology, Rigshospitalet, Copenhagen, Denmark \AND
\firstname Claes N. \surname Ladefoged \email{claes.noehr.ladefoged@regionh.dk} \AND
\addr Department of Clinical Physiology and Nuclear Medicine, Rigshospitalet, Copenhagen, Denmark \AND
\addr Department of Applied Mathematics and Computer Science, Technical University of Denmark, Kongens Lyngby, Denmark \AND
\name Desmond Elliott \email{de@di.ku.dk} \AND
\addr Department of Computer Science, University of Copenhagen, Copenhagen, Denmark \AND
\firstname Gitte M. \surname Knudsen \email{gmk@nru.dk} \AND
\addr Department of Clinical Medicine, University of Copenhagen, Copenhagen, Denmark \AND
\firstname Lars H. \surname Pinborg \email{lars.pinborg@nru.dk}
\AND
\addr Epilepsy Clinic, Department of Neurology, Rigshospitalet, Copenhagen, Denmark \AND
\addr Neurobiology Research Unit, Rigshospitalet and Department of Clinical Medicine, University of Copenhagen, Copenhagen, Denmark \AND
\name Melanie Ganz \email{melanie.ganz@nru.dk} \AND
\addr Neurobiology Research Unit, Rigshospitalet, Copenhagen, Denmark \AND
\addr Department of Computer Science, University of Copenhagen, Copenhagen, Denmark
}

\begin{document}

\maketitle

\begin{abstract}
    Natural language processing (NLP) in the medical domain can underperform in real-world applications involving small datasets in a non-English language with few labeled samples and imbalanced classes. There is yet no consensus on how to approach this problem. We evaluated a set of NLP models including BERT-like transformers, few-shot learning with sentence transformers (SetFit), and prompted large language models (LLM), using three datasets of radiology reports on magnetic resonance images of epilepsy patients in Danish, a low-resource language. Our results indicate that BERT-like models pretrained in the target domain of radiology reports currently offer the optimal performances for this scenario. Notably, the SetFit and LLM models underperformed compared to BERT-like models, with LLM performing the worst. Importantly, none of the models investigated was sufficiently accurate to allow for text classification without any supervision. However, they show potential for data filtering, which could reduce the amount of manual labeling required. \looseness=-1
\end{abstract}


\section{Introduction}

The increasing access to electronic health records (EHR) has opened unparalleled opportunities for the processing of big data in the medical domain. However, the information contained in EHR is largely unstructured or semi-structured, and further processing is required to obtain the desired information. In this context, a prominent recurring task is the extraction of relevant labels from medical texts associated with external data. This is particularly relevant in radiology where clinical findings present in images can be extracted from the matching reports written by radiologists. These features can then be used in correspondence with the images, for example when creating labeled image data sets for image classification tasks. Labeling medical reports can be very time-consuming and, depending on the context, substantial efforts may be required even to create relatively small datasets. Furthermore, many pathologies have a low prevalence and will result in datasets with highly imbalanced classes. On a large scale, manually performing this type of labeling task is intractable, and automated methods are therefore required.

Practical applications of natural language processing (NLP) in the medical domain can suffer from compounded issues, including non-English language, a small number of labeled samples, and class imbalance. These factors can all adversely impact the performance of NLP models in unique ways and a reliable approach to jointly tackle these issues is yet to be determined. In this work, we focus on a realistic use case of labeling radiology reports of magnetic resonance images (MRI) in the Danish language in a cohort of epilepsy patients. Our primary goal is to evaluate the current state-of-the-art of NLP models in this context and provide a comparative baseline for researchers with similar tasks.

\section{Related Works}

The usefulness of NLP to extract information from radiological text is increasingly recognized and specialized models such as RadBERT  \citep{yan2022radbert} have been proposed as a general approach in the English language. Work utilizing these models has, for example, been applied to radiological descriptions of MRI and their results suggest that the automated large scale labeling of radiology reports in English is achievable with high accuracy \citep{wood_automated_2020}. 

To our knowledge, the application of NLP models to radiology reports beyond the English language, and especially in low-resource languages, remains limited. BERT-like models for text summarization in Japanese \citep{nishio2024fully} and multilingual support (English, Portuguese, and German) \citep{lindo2023multilingual} have been suggested to provide adequate performance. However, models for text classification in Polish \citep{obuchowski_information_2023}, French, and German \citep{mottin2023multilingual}, have all shown reduced performance compared to their English counterpart. Despite their generally superior performance, large language models (LLM) have seen little application for radiological text in non-English language. Recent work by \cite{matsuo2024exploring} investigating the classification of radiological text in Japanese suggests that translating the text to English improves classification accuracy with a multilingual LLM (GPT3.5). We are not aware of published studies on radiology reports in a non-English language using sentence transformers.

\section{Methods}

\subsection{Dataset}

A dataset of 16,899 MRI reports in the Danish language describing the brain scans of 4,769 patients with ICD-10 code G40* (epilepsy) was obtained. Example of short and long MRI reports for epilepsy patients are given in Figures and \ref{methods:short_report} and \ref{methods:long_report}. Additionally, a corpus of 1,2 million radiology reports in Danish were retrieved in bulk, irrespective of modality (MRI, computed tomography, X-ray, ultrasound), body parts, and disease, and used for pretraining the BERT-like models (see section \ref{section:bert}). All radiology reports were retrieved from a centralized picture archiving and communication system at Rigshospitalet in the Capital Region of Denmark, and covered the period 2017-2022. This study was approved by the National Scientific Ethics Committee of Denmark [D1936897].

Three types of abnormalities relevant to epilepsy were labeled in the MRI reports of epilepsy patients: focal cortical dysplasia (FCD) (n=1,122), mesial temporal sclerosis (MTS) (n=904), and hippocampal abnormalities (HA) (n=992). Reports with mention of the abnormalities were identified using regular expressions and manually labeled by a medical student (HK) under the supervision of an expert neurologist (LHP). The FCD dataset was also labeled by a second clinician (MP) to investigate inter-rater agreement. The FCD and MTS datasets represent cases where the radiologist described the presence or absence of a pathology directly, and is often associated with a degree of certainty. To account for the variable degree of confidence, the prefixes negative, probable, highly probable (only for FCD), and positive were manually appended to the FCD (n=877/86/93/66) and MTS (n=668/104/132) labels. The HA dataset present more complex cases where abnormal hippocampal features (e.g., atrophy, hyperintensities) are described, but a  pathological diagnosis is not explicitly indicated (see Appendix \ref{appendix:sentences} for an example). In this case, reports were summarily labeled as abnormal if any type of abnormality was present, and normal otherwise. The HA dataset contained n=267/725 normal and abnormal labels. Labeling of the FCD, MTS, and HA datasets took approximately 35, 25, and 30 hours, respectively. Training and test sets were created for each datasets using 80\%/20\% splits, and 20\% of the training data was used for validation. An overview of the data extraction and labeling is presented in Figure \ref{fig:overview}.

\begin{figure}
\begin{framed}
\begin{minipage}{38em}
    {\setlength{\parindent}{0cm} \footnotesize
    \textbf{Original text in Danish:}\\
    
    Undersøgelse: MR cerebrum uden kontrast\\
    
    Indikation: mr af cerebrum i generel anæstesi som kontrol af focal cortical dysplasi\\
    
    Beskrivelse: MR-skanning af cerebrum viser sammenlignet med skanningen fra den [date] til [date] en fokal forandring lateralt i venstre frontallap  opfattes som fokal kortikal dysplasi. Herudover ses der uændrede lette hvid substansforandringer periventrikulært posteriort bilateralt samt en diskret lille hvid substansforandringer i venstre corona radiata / centralt. Der er ikke nytilkomne fokale forandringer.\\
    
    Konklusion:  Uændret kortikal læsion frontalt venstre side opfattes som kortikal dysplasi Hvid substansforandringer se tekst.\\
    
    \textbf{Translation to English:}\\
    
    Examination: MR cerebrum without contrast\\
    
    Indication: MRI of the cerebrum under general anesthesia to check for focal cortical dysplasia\\
    
    Description: MR scan of the cerebrum shows, compared to the scan from [date] to [date], a focal change laterally in the left frontal lobe perceived as focal cortical dysplasia. In addition, there are unchanged light white matter changes periventricularly posteriorly bilaterally and a discreet small white matter change in the left corona radiata / centrally. There are no new focal changes.\\
    
    Conclusion: Unchanged cortical lesion frontal left side is perceived as cortical dysplasia White matter changes see text.\\
    }
\end{minipage}
\end{framed}
\caption{An example short radiology reports describing a patient with focal cortical dysplasia (FCD). Dates were anonymized for presentation purposes.}
\label{methods:short_report}
\end{figure}

\subsection{Preprocessing}

To reduce the complexity of the radiology reports and provide information more relevant to the classification task, only sentences containing selected regular expressions related to the target labels were kept. Every text was divided into individual sentences using the Danish NLP framework \texttt{DaCy} (v2.7.7, da\_dacy\_large\_trf-0.2.0) \citep{enevoldsen2021dacy} and relevant sentences were identified using the same regular expression which was used to initially identify the radiology reports. The selected sentences were then concatenated and used as input for the classification models. An overview of the preprocessing is presented in Figure \ref{fig:overview} and examples of preprocessed sentences for some of the labeled categories are given in Appendix \ref{appendix:sentences}.

\begin{figure}
    \centering
    \includegraphics[width=\linewidth]{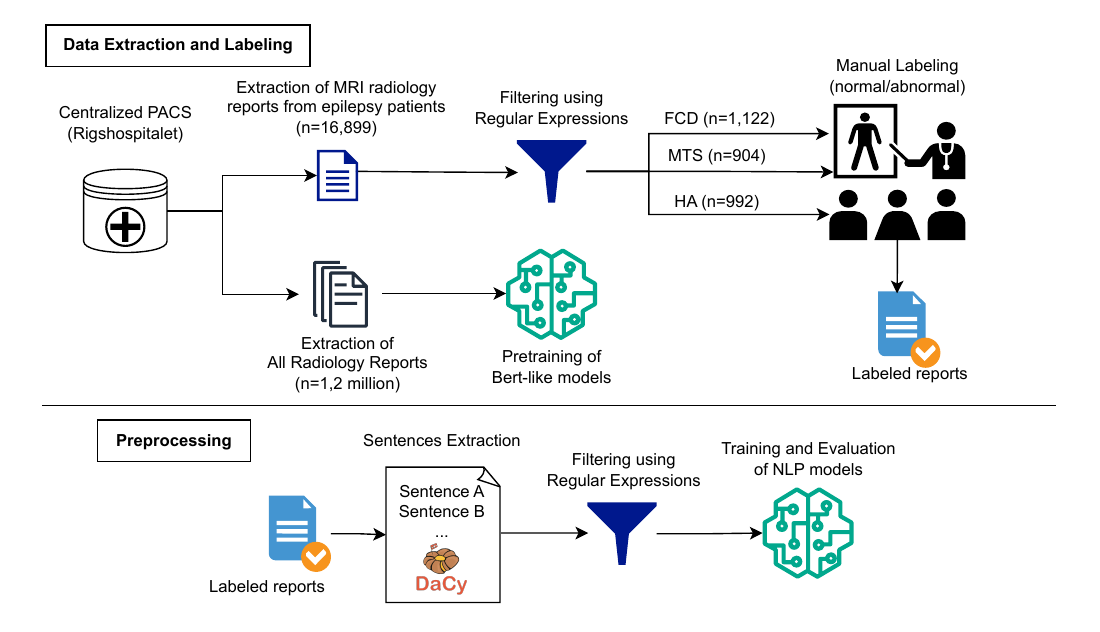}
    \caption{Overview of the data extraction, labeling and preprocessing. FCD: focal cortical dysplasia, MTS: mesial temporal sclerosis, HA: hippocalpal abnormality, PACS: picture archiving and communication system.}
    \label{fig:overview}
\end{figure}

\subsection{Natural Language Processing Models}

Three approaches were evaluated: (BERT-like) transformers, few-shot learning with sentence transformers (SetFit), and LLM. The NLP models were trained and evaluated using the \texttt{transformers} package (v4.40.1) from Huggingface. Details on hyperparameter optimization are available in Appendix \ref{appendix:optimization}.

\subsubsection{BERT-like transformer models \label{section:bert}}

BERT-like transformer models natively supporting the Danish language (RøBÆRTa\footnote{https://huggingface.co/DDSC/roberta-base-danish}) and multilingual text including Danish (XLM-RoBERTa) \citep{DBLP:journals/corr/abs-1911-02116} were evaluated. These checkpoint models were used with and without continued pretraining~\citep{gururangan-etal-2020-dont} on the corpus of 1,2 millions radiology reports. Model pretraining was performed using whole-word masking. Fine-tuning for text classification was performed using a sequence classification head with weighted (binary) cross-entropy loss. Fine-tuning was performed over 50 epochs with a batch size of 16.

\subsubsection{SetFit}

The SetFit approach (setfit, v1.0.3) \citep{tunstall_efficient_2022} was evaluated using a sentence transformer model for Danish (dfm-sentence-encoder-large\footnote{https://huggingface.co/KennethEnevoldsen/dfm-sentence-encoder-large-exp2-no-lang-align}) and a model with multilingual support (distiluse-base-multilingual-cased-v2\footnote{https://huggingface.co/sentence-transformers/distiluse-base-multilingual-cased-v2}) \citep{reimers-2020-multilingual-sentence-bert}. Training was performed by first pretraining the model's body for 25,000 steps and then fine-tuning the model end-to-end (i.e., including the classification head) for 50 epochs. A differentiable classification head using a linear layer to map the embeddings to the classes was used. In all cases, a batch size of 8 was used.

\subsubsection{Large Language Models}

Three different LLMs were evaluated: a Danish LLM (munin-neuralbeagle-7b\footnote{https://huggingface.co/RJuro/munin-neuralbeagle-7b}), a general purpose LLM primarily trained for the English language (Meta-Llama-3-70B-Instruct\footnote{https://huggingface.co/meta-llama/Meta-Llama-3-70B-Instruct}) \citep{llama3modelcard}, and a LLM tailored to the health domain in English (BioMistral-7B\footnote{https://huggingface.co/BioMistral/BioMistral-7B}) \citep{labrak2024biomistral}. LLMs were evaluated using few-shots prompting. For the \texttt{Meta-\allowbreak Llama-\allowbreak 3-\allowbreak 70B-\allowbreak Instruct} and the \texttt{BioMistral-7B} models, the texts were translated from Danish to English using the MADLAD-400-10B-MT model\footnote{https://huggingface.co/google/madlad400-10b-mt} \citep{kudugunta2023madlad400}. For each model, we followed the prompt formatting for few-shot inference as recommended by the model's developers. The corresponding prompt templates are given in Appendix \ref{appendix:prompts}. 

\section{Results}

The agreement (Cohen's kappa) between the two raters for the FCD dataset was 0.83. Table \ref{tab:results} presents the evaluation metrics for the classifiers on the different datasets.

The performance of each model is presented in Table \ref{tab:results}. Across our three datasets, BERT-like models displayed the highest performance, with \texttt{DanskBERT (pretrained)} ranking first for the FCD and MTS datasets and \texttt{xlm-roberta-base (pretrained)} for the HA dataset. Expectedly, in almost all cases pretraining the BERT-like models on the corpus of 1,2 million radiology reports improved the predictive performances of the models, with the notable exception of \texttt{xlm-roberta-base} on the FCD dataset. Overall, both the SetFit and LLM models displayed comparatively reduced performances, with the LLMs ranking among the worst models.

\begin{figure}[h]
    \centering
    \begin{minipage}[t]{0.32\textwidth}
        \centering
        \includegraphics[width=0.7\textwidth]{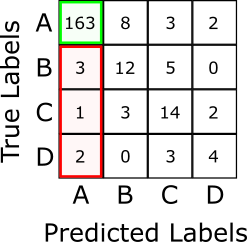}\\
        (A) xlm-roberta-base (original)
    \end{minipage}
    \hfill
    \begin{minipage}[t]{0.32\textwidth}
        \centering
        \includegraphics[width=0.7\textwidth]{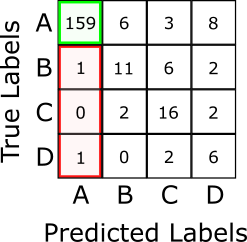} \\
        (B) xlm-roberta-base (pretrained)
    \end{minipage}
    \hfill
    \begin{minipage}[t]{0.32\textwidth}
        \centering
        \includegraphics[width=0.7\textwidth]{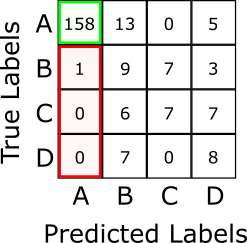} \\
        (C) distiluse-base-multilingual-cased-v2
    \end{minipage}
    \label{fig:confusion}
    \caption{Confusion matrices of selected classifiers on the FCD test dataset. Recall is \crule[green]/(\crule[green]+\crule[red]). A: No FCD, B: Potential FCD, C: Highly Probable FCD, D: FCD}    
\end{figure}

Figure \ref{fig:confusion} shows examples of confusion matrices of selected models for the FCD dataset.

\begin{table}
    \label{tab:results}
    {\caption{Evaluation metrics of the classifiers. FCD: focal cortical dysplasia, MTS: mesial temporal sclerosis, HA: hippocampal abnormalities. The highest metric for each dataset is marked in bold.}}%
    {\begin{tabular}{p{74mm}|c|c|c|c|c|c}
    \hline
    & \multicolumn{3}{c|}{F1-score (macro)} & \multicolumn{3}{c}{Balanced Accuracy} \\
    \hline
    \bfseries Model & \bfseries FCD & \bfseries MTS & \bfseries HA &  \bfseries FCD & \bfseries MTS & \bfseries HA \\
    \hline
    roberta-base-danish (original) & 0.60 & 0.76 & 0.69 & 0.65 & 0.78 & 0.72 \\
    roberta-base-danish (pretrained) & 0.62 & 0.83 & 0.70 & 0.66 & 0.85 & 0.72 \\
    DanskBERT (original) & 0.75 & 0.81 & 0.67 & 0.77 & 0.81 & 0.72 \\
    DanskBERT (pretrained) & \textbf{0.75} & \textbf{0.88} & 0.70 & \textbf{0.81} & \textbf{0.91} & 0.71 \\
    xlm-roberta-base (original) & 0.73 & 0.79 & 0.70 & 0.76 & 0.79 & 0.73 \\
    xlm-roberta-base (pretrained) & 0.69 & 0.85 & \textbf{0.71} & 0.69 & 0.86 & \textbf{0.74} \\
    \hline
    distiluse-base-multilingual-cased-v2 & 0.46 & 0.79 & 0.68 & 0.49 & 0.80 & 0.71 \\
    dfm-sentence-encoder-large & 0.66 & 0.79 & 0.65 & 0.71 & 0.81 & 0.65 \\
    \hline
    munin-neuralbeagle-7b & 0.45 & 0.71 & 0.65 & 0.55 & 0.72 & 0.70 \\
    Meta-Llama-3-70B-Instruct (w/ translation) & 0.53 & 0.62 & 0.69 & 0.65 & 0.69 & 0.74 \\
    BioMistral-7b (w/ translation) & 0.38 & 0.47 & 0.56 & 0.55 & 0.57 & 0.56
    \end{tabular}}
\end{table}

\section{Discussion and Conclusion}

In this work, we evaluated a range of approaches and datasets representing the task of medical text classification in small and imbalanced datasets in a non-English Language.

Despite the recent rise of LLMs in NLP, most applications in non-English radiology reports have focused on using BERT-like transformer models. In our evaluation, all best-performing models were in fact BERT-like models, indicating that these simpler models continue to deliver state-of-the-art performances in targeted applications. No single model outperformed all others, however, the pretrained DanskBERT model did provide the best performance for the FCD and MTS datasets, and competitive performances for the HA dataset, suggesting that this model may be best suited to our task. 

Pretraining on a domain-specific corpus has consistently been shown to improve various NLP tasks. A popular example of this is BioBERT which has gained popularity in the scientific domain \citep{lee2020biobert}. Generally, the availability of transformers pretrained for specific domains in non-English languages is a core issue for the generalizability of NLP approaches. Here, we evaluated our models with and without pretraining to provide a point of reference showcasing the possible gain in performance. As expected, pretraining did improve performance in almost all cases. However, it is important to emphasize that obtaining a relevant corpus may be non-trivial and can require substantial time and/or resources. For example, the corpus of 1,2 million radiology reports used in our work took approximately 1.5 years to extract due to limitations of the hospital's IT infrastructure. Although automated, in a time-limited project the duration of a similar process would need to be carefully weighed against the potential gain in performance.   

The SetFit approach, which optimizes the embeddings of sentence transformers \citep{reimers-2019-sentence-bert}, has been introduced as a competitive approach to the BERT-like transformers for small datasets. Contrary to our expectations, this approach rarely outperformed the BERT-like transformers, with and without pretraining. This may be due to the fact that a sentence in a radiology report may contain many details that are not directly relevant to the classification task, therefore leading to sentence embeddings that may be inadequate for isolating specific information. However, more research on this topic would be required to disentangle this issue.

Large language models performed poorly in our evaluation. With 7 billion parameters \texttt{munin-neuralbeagle-7b} is a relatively small LLM, but it is ranked among the top models on the Mainland Scandinavian NLG leaderboard\footnote{https://scandeval.com/mainland-scandinavian-nlg}. \texttt{BioMistral-7B} is one of the latest generation of LLM adapted to the medical domain. The \texttt{Meta-Llama-3-70B-Instruct} model is by far the largest model included in this study and has exhibited state-of-the-art performances in a wide range of NLP tasks \citep{llama3modelcard}. Although we have used the recommended approaches for generating few-shot prompts, a different strategy may yield better results. Furthermore, the translation from Danish to English was in a few cases suboptimal, which may have negatively impacted the predictions. Investigating the quality of different translation models may potentially lead to improved performance for the LLMs. Overall, the poor performance of the LLMs in our scenario is surprising and warrants further research in this topic. 

The performance required from an NLP models is strongly dependent on the downstream application. Although there is no definite agreement, an accuracy equivalent or superior to that of an expert clinician, which has been shown in similar work to be above 90\% \citep{wood_automated_2020}, is often desired. It is therefore important to emphasize that, under this expectation, none of the models evaluated in our setting exhibited performances sufficient to provide a reliable and fully automated solution. However, a closer look at the confusion matrices reveals that some of the classifiers have an almost perfect recall for the most numerous class (Fig. \ref{fig:confusion}B-C). Therefore, when manually labeling large datasets a substantial amount of work could potentially be avoided by first using the classifier to identify the reports belonging to that class and then only processing the remainder. However, the performance of this approach is heavily dependent on the dataset and would have to be carefully validated in each case.


\acks{This work was supported by the Lundbeck Foundation (grant R279-2018-1145, BrainDrugs).}

%
\ethics{The work follows appropriate ethical standards in conducting research and writing the manuscript, following all applicable laws and regulations regarding treatment of animals or human subjects.}

\coi{We declare we don't have conflicts of interest.}

\data{The data used in this study contains personal information and therefore underlies GDPR. Therefore, it cannot be shared openly, but a request to share it securely under a data usage agreement can be made. The code used for this project is openly available at \url{https://github.com/vbeliveau/radiology-text-classification}}

\bibliography{references}

\clearpage
\appendix

\section{Example of a long radiology report}

\begin{figure}[hb]
\begin{framed}
\begin{minipage}{38em}
{\setlength{\parindent}{0cm} \footnotesize
    \textbf{Original text:}\\
    
    Undersøgelse: MR cerebrum uden og med kontrast\\

    Indikation: kortical dysplasi/lavgradsgliom fulgt siden [year].\\

    Beskrivelse: MR cerebrum uden og med i.v. kontrast, sammenholdt med undersøgelse fra den [dato], ses der uændret størrelse og udseende af T2 / FLAIR hyperintense forandringer kortikosubkortikalt i venstre parietallap med enkelte foci af kontrastopladning. Ligeledes uændret lineær T2 / FLAIR hyperintensitet gående fra den ovennævnte forandring til baghorn af venstre lateralventrikel (så kaldt ”transmantle sign”). Derudover ses der enkelte foci af T2 / FLAIR hyperintensitet i bilaterale periventrikulære og frontale subkortikale hvid substans, uspecifikke, uændret siden sidste. Ingen nytilkomne forandringer. Der er ingen friske infarkter eller blødninger. Ingen ekstraaksiale ansamlinger eller hydrocephalus. Frie basale cisterner. Ingen nytilkomne patologiske kontrastopladninger. Minimal slimhindefortykkelse inferiort i bilaterale sinus maxillaris. Medskannet kalvariet, orbitae og paranasale sinus er i øvrigt upåfaldende. Ingen patologiske signaler fra mastoidceller.\\

    Konklusion: Uændret T2 / FLAIR hyperintensitet kortikosubkortikalt i venstre parietallap med transmantle tegn, mest foreneligt med fokale kortikal dysplasi, alternativt/ mindre sandsynlig differentialdiagnose er lav grad gliom. Intet nytilkommet siden sidste- se venligst tekst.\\
    
    \textbf{Translation:}\\
    
    Examination: MR cerebrum without and with contrast\\
    
    Indication: cortical dysplasia/low-grade glioma followed since [date].\\
    
    Description: MR cerebrum without and with i.v. contrast, compared with examination from the [date], unchanged size and appearance of T2 / FLAIR hyperintense corticosubcortical changes in the left parietal lobe with single foci of contrast loading are seen. Likewise, unchanged linear T2 / FLAIR hyperintensity going from the above-mentioned change to the posterior horn of the left lateral ventricle (so-called "transmantle sign"). In addition, single foci of T2 / FLAIR hyperintensity are seen in bilateral periventricular and frontal subcortical white matter, non-specific, unchanged since last. No new changes. There are no fresh infarcts or bleeding. No extraaxial collections or hydrocephalus. Free basal cisterns. No new pathologic contrast charges. Minimal mucosal thickening inferiorly in bilateral maxillary sinuses. Scanned calvaria, orbitae and paranasal sinuses are otherwise unremarkable. No pathological signals from mastoid cells.\\
    
    Conclusion: Unchanged T2 / FLAIR hyperintensity corticosubcortically in the left parietal lobe with transmantle signs, most compatible with focal cortical dysplasia, alternative/ less probable differential diagnosis is low grade glioma. Nothing new since last - please see text.
}\end{minipage}
\end{framed}
\caption{An example long radiology reports describing a patient with focal cortical dysplasia (FCD). Dates were anonymized for presentation purposes.}
\label{methods:long_report}
\end{figure}

\newpage

\section{Methods}

    \subsection{Example Preprocessed Sentences \label{appendix:sentences}}

    \begin{framed}
    {\setlength{\parindent}{0cm}    
    \textbf{No FCD}\\
    Ingen tegn på mesial temporalsklerose, heterotopi eller fokal kortikal dysplasi.\\
    \textit{(Translation) No evidence of mesial temporal sclerosis, heterotopia or focal cortical dysplasia.}

    \noindent\rule{\textwidth}{1pt}
    
    \textbf{FCD}\\
    Fund forenelig med fokal kortikal dysplasi. Derudover ses ingen andre tegn på kortikale dysplasier eller heterotopier. MRD: Tegn på venstresidig transmantle kortikal dysplasi.\\
    \textit{(Translation) Findings consistent with focal cortical dysplasia, no other evidence of cortical dysplasia or heterotopia. MRD: Signs of left-sided transmantle cortical dysplasia.}
    
    \noindent\rule{\textwidth}{1pt}
    
    \textbf{No MTS}\\
    Derudover lille ependymal cyste over caput af venstre hippocampus, men ingen tegn på  MTS.\\
    \textit{(Translation) In addition, small ependymal cyst over caput of left hippocampus, but no evidence of MTS.}
    
    \noindent\rule{\textwidth}{1pt}
    
    \textbf{MTS}\\
    Der ses tegn på mesial temporal sklerose på begge sider. Sklerosen involverer alle de tre del af hippocampi. Konklusion: Bilateral mesial temporal sklerose.\\
    \textit{(Translation) Signs of mesial temporal sclerosis are present on both sides, and the sclerosis involves all three parts of the hippocampus. Conclusion: Bilateral mesial temporal sclerosis.}

    \noindent\rule{\textwidth}{1pt}
    
    \textbf{No HA}\\
    Hippocampi fremstår symmetriske med volumen, signalintensitet og arkitektur indenfor det normale.\\
    \textit{(Translation) Hippocampus appears symmetrical with volume, signal intensity and architecture within normal range.}

    \noindent\rule{\textwidth}{1pt}
    
    \textbf{HA}\\
    Der er tilkommet atrofi af hele venstre hemisfære, hovedsageligt med tab af grå substans og atrofi af hippocampus.\\
    \textit{(Translation) There is atrophy of the entire left hemisphere, mainly with loss of gray matter and atrophy of the hippocampus.}
    }
    \end{framed}

    \subsection{Hyperparameters Optimization \label{appendix:optimization}}

    For all models, the BERT-like models and the SetFit models, hyperparameter optimization was performed using Optuna (v3.6.1). In a all cases, 100 trials were used to select the optimal hyperparameters and 80\% of the original training data was used for training and 20\% for validation. F$_1$-score was used as target metric to optimize classification.
    
    The \texttt{learning\_rate} was optimized in the range $[1e{-7}, 1e{-5}]$ for \texttt{roberta-base-danish}, and in the range $[1e{-6}, 1e{-4}]$ for \texttt{DanskBERT} and \texttt{xml-roberta-base}. For all models, \texttt{weight\_decay} was optimized in the range $[1e{-4}, 1e{-2}]$.

    For the SetFit models, the embeddings of the models were first optimized for each individual datasets by training the model body with \texttt{body\_learning\_rate} (range $[5e^{-7}, 5e^{-6}]$) being optimized. The models were then trained end-to-end, including the classification head, by optimizing the parameters \texttt{body\_learning\_rate} (range $[5e^{-7}, 5e^{-6}]$), \texttt{head\_learning\_rate} (range $[1e^{-4}, 1e^{-1}]$), \texttt{l2\_weight} (range $[1e^{-4}, 1e^{-3}]$). There were two exceptions to this process. Firstly, for the \texttt{distiluse-base-multilingual-cased-v2} applied to the FCD dataset, \texttt{body\_learning\_rate} was optimized range $[5e^{-7}, 5e^{-5}]$ and \texttt{head\_learning\_rate} in the range $[1e^{-5}, 1e^{-3}]$.  Secondly, the \texttt{dfm-sentence-encoder-large-exp2-no-lang-align} model applied to the HA dataset where training the model body did not improve the related validation loss. In that case, this step was skipped.
    
    In the case of LLMs, the models were not further trained and, therefore, the only hyperparameter to optimize is the number of shots to be included during inference. However, as the inference time is in the range of 2-3 minutes per sample for the larger LLMs in our setup, this optimization was performed only for the smallest model. For each dataset, the optimal number of shots for \texttt{munin-neuralbeagle-7b} was selected in the range [1, 7] using 20\% of the training data for validation and F$_1$-score as target metric. For the other two LLMs, the number of shots was set to 10.

    \newpage

    \subsection{Templates for LLM prompts 		\label{appendix:prompts}}

    \subsubsection*{munin-neuralbeagle-7b}

    \begin{framed}
    {\setlength{\parindent}{0cm}
    You are an experienced radiologist that help users extract information from radiology reports. Categorize the text in $<<<>>>$ into one of the following predefined categories:\\

    LABEL 1\\
    ...\\
    LABEL N\\
    
    You will only respond with the category. Do not include the word "Category". Do not provide explanations or notes.\\
    
    \#\#\#\#\\
    Here are some examples:\\
    
    Inquiry: SAMPLE TEXT 1\\
    Category: LABEL 1\\
    Inquiry: SAMPLE TEXT N\\
    Category: LABEL N\\
    \#\#\#\\
    
    $<<<$\\
    Inquiry: QUERY TEXT\\
    $>>>$
    }
    \end{framed}
    
    \subsubsection*{BioMistral-7b}

    \begin{framed}
    {\setlength{\parindent}{0cm}
    $<$s$>$$[$INST$]$You are an experienced radiologist that help users extract infromation from radiology reports. Your task is to categorize texts in the following categories:\\
    
    LABEL 1\\
    ...\\
    LABEL N\\
    
    You will only respond with the category. Do not include the word "Category". Do not provide explanations or notes.\\
    Categorize the text: SAMPLE TEXT 1\\
    $[$/INST$]$LABEL 1$<$/s$>$$[$INST$]$\\
    Categorize the text: SAMPLE TEXT N\\
    $[$/INST$]$LABEL N$<$/s$>$$[$INST$]$\\
    Categorize the text: QUERY TEXT\\
    $[$/INST$]$
    }
    \end{framed}

    \subsubsection*{Meta-Llama-3-70B-Instruct}

    \begin{framed}
    {\setlength{\parindent}{0cm}
    [\{'role': 'system',\\
    'content': 'You are an experienced radiologist that help users extract information from radiology reports. Your task is to categorize texts in the following categories:
    LABEL 1, ..., LABEL N. You will only respond with the category. Do not include the word "Category". Do not provide explanations or notes.'\},\\
    \{'role': 'user', 'content': 'Categorize the text: SAMPLE TEXT 1'\},\\
    \{'role': 'assistant', 'content': 'LABEL 1'\},\\
    \{'role': 'user', 'content': 'Categorize the text: SAMPLE TEXT N'\},\\
    \{'role': 'assistant', 'content': 'LABEL N'\},\\
    \{'role': 'user', 'content': 'Categorize the text: QUERY TEXT'\}]
    }
    \end{framed}

\end{document}